\documentclass[journal]{vgtc}                

\ifpdf
  \pdfoutput=1\relax                   
  \usepackage{graphicx}                
  \DeclareGraphicsExtensions{.pdf,.png,.jpg,.jpeg} 
\else
  \usepackage{graphicx}                
  \DeclareGraphicsExtensions{.eps}     
\fi%

\graphicspath{{figures/}{pictures/}{images/}{./}} 

\usepackage{microtype}                 
\PassOptionsToPackage{warn}{textcomp}  
\usepackage{textcomp}                  
\usepackage{mathptmx}                  
\usepackage{times}                     
\usepackage{cite}                      
\usepackage{listings}

\usepackage{xspace,xpunctuate}

\newcommand{\ie}{{i.e.,}\xspace}
\newcommand{\cf}{{cf.}\xspace}
\newcommand{\eg}{{e.g.,}\xspace}
\newcommand{\ea}{{et~al\xperiod}\xspace}

\onlineid{1126}

\vgtccategory{Research}
\vgtcpapertype{algorithm/technique}

\title{Neural Activation Patterns (NAPs):\\Visual Explainability of Learned Concepts}

\author{Alex Bäuerle\thanks{Both authors contributed equally to this research.}, Daniel Jönsson\footnotemark[1], and Timo Ropinski}
\authorfooter{
\item
 Daniel Jönsson is with Ulm University and Linköping University. Email: daniel.jonsson@liu.se.
\item
 Alex Bäuerle and Timo Ropinski are with Ulm University. Email: see \url{https://a13x.io}
}

\shortauthortitle{Bäuerle and Jönsson \MakeLowercase{\textit{et al.}}: Neural Activation Patterns}

\abstract{
A key to deciphering the inner workings of neural networks is understanding what a model has learned. 
Promising methods for discovering learned features are based on analyzing activation values, whereby current techniques focus on analyzing high activation values to reveal interesting features on a neuron level.
However, analyzing high activation values limits layer-level concept discovery.
We present a method that instead takes into account the entire activation distribution.
By extracting similar activation profiles within the high-dimensional activation space of a neural network layer, we find groups of inputs that are treated similarly.
These input groups represent neural activation patterns (NAPs) and can be used to visualize and interpret learned layer concepts.
We release a framework with which NAPs can be extracted from pre-trained models and provide a visual introspection tool that can be used to analyze NAPs.
We tested our method with a variety of networks and show how it complements existing methods for analyzing neural network activation values.
} 


\CCScatlist{ 
 \CCScat{K.6.1}{Management of Computing and Information Systems}%
{Project and People Management}{Life Cycle};
 \CCScat{K.7.m}{The Computing Profession}{Miscellaneous}{Ethics}
}

\teaser{
  \centering
  \includegraphics[width=\linewidth]{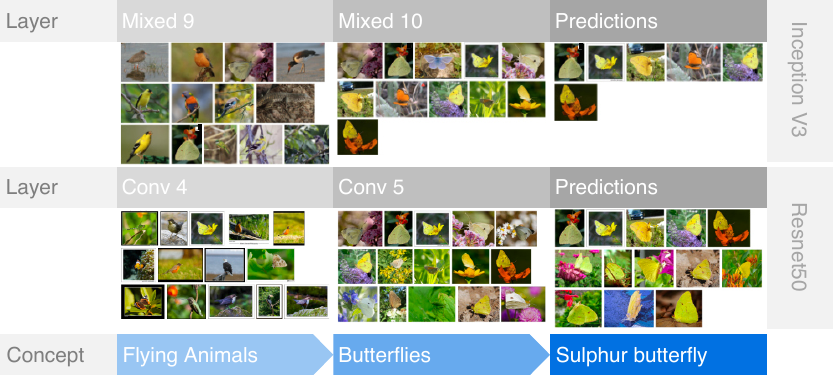}
  \caption{
  Examples of discovered concepts across layers and networks.
  The concept of a sulphur butterfly is built based on different more high-level concepts in previous layers.
  For this concept, Inception V3 and Resnet50 show similar NAPs.
  }
	\label{fig:teaser}
}

\vgtcinsertpkg
\usepackage{caption}
\usepackage{subcaption}

\begin{document}


\firstsection{Introduction}

\maketitle

Advances in the development and utilization of deep neural networks have been significant in recent years.
As the utilization of these networks expands, building a fundamental understanding of what they have learned also becomes evermore important~\cite{adebayo2020debugging}.
The discovery and analysis of human interpretable concepts in neural networks can not only serve as a way of detecting errors, improving accuracy, or surfacing bias~\cite{olah2018building, kim2018interpretability}, it can also be essential for discovering and communicating the decision processes occurring inside a neural network.
Given the importance of explainable AI to support these understanding and debugging processes, further advancing these techniques is imperative for building reliable, accurate, and explainable AI systems.

Visualization of neural network activations is an important aspect of many explainability approaches~\cite{kahng2017cti,hohman2019s}.
Perhaps the most common technique to visualize what a network has learned is to extract and display the highest activating inputs for individual units (components of the layer output activation vector, also referred to as neurons, filters, or features in other works)~\cite{fong2018net2vec}.
These highly activating examples can illustrate what the units under investigation have learned.
Similarly, Feature Visualization can obtain an input that maximizes the unit activation; illustrating which features excite the unit under investigation~\cite{olah2017feature}.

\begin{figure}[t]
\centering
\includegraphics[width=\linewidth]{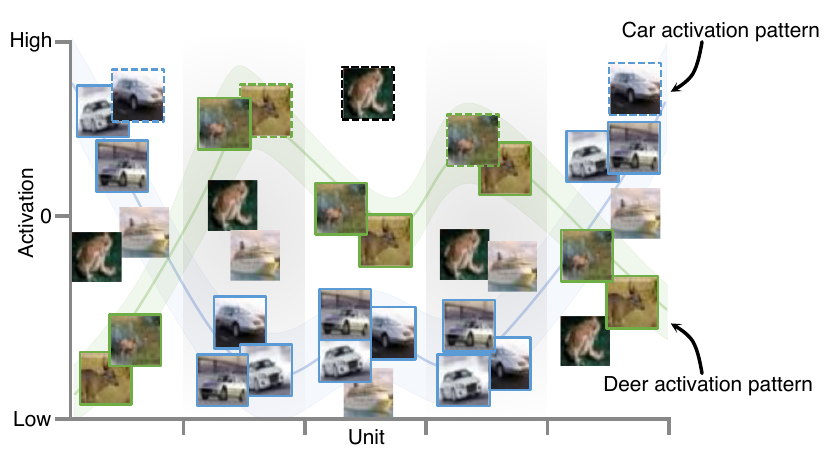}
\caption{Illustration of neural network layer output activation per unit (horizontal axis) for six different image inputs. 
Max activation extracts the highest activating inputs (dotted border).
Our neural activation patterns, indicated in blue and green, instead consider inputs with similar activation profiles over all units and thus produce a more complete picture of what the layer learned. 
}
\label{fig:ActivationDistribution}
\end{figure}

Visualizing what individual units have learned is an important clue towards a better understanding of trained neural networks.
However, each neural network layer extracts many features that, in combination, represent which concepts a layer has learned (\cf{} \autoref{fig:ActivationDistribution}).
The discovery of layer-level concepts goes beyond inspecting individual units because a layer can make use of combined unit activations.
Thus, to investigate neural network layers in their entirety, it is necessary to take the activation of all layer units into account.
Focusing only on the highest activation values ignores combinations within the high-dimensional layer activation output vector.
Therefore, these operations fail to capture the complex concepts that are needed to provide expressive explanations on a layer level.

In this work, we follow the notion that \emph{representations are distributed, and thus filters must be studied in conjunction}~\cite{fong2018net2vec}.
Therefore, we treat the activations produced by a set of inputs as a distribution over the units that span the activation space of the layer. 
Instead of extracting the maximum activation in this distribution, we find combinations of unit activations that respond similarly to different inputs within the distribution.
The inputs treated similarly by a layer will thus have a specific activation profile, which we refer to as a Neural Activation Pattern (NAP), whereby the inputs associated with a NAP are used as a visual means for the interpretation of learned concepts.
Because NAPs consider combinations of features, it is possible to discover concepts learned on a layer level, and thus NAPs complement neural network visualization building blocks working on the individual unit level.

We demonstrate the capabilities of NAPs on various deep neural networks, ranging from simplistic networks trained on MNIST~\cite{lecun2010mnist} to complex networks such as Inception V3~\cite{szegedy2016rethinking} trained on ImageNet~\cite{deng2009imagenet}.
Furthermore, we present a visual environment designed for inspecting and comparing NAPs.
The visual environment depicts different views of the NAPs in the form of their corresponding inputs, predictions, and labels, as well as their activation distributions.
We show that interpretable layer-level learned concepts can be extracted from neural networks using these views. 
NAPs can thus contribute to building a fundamental understanding of what neural network layers have learned.
Furthermore, by comparing NAPs of different networks, it becomes possible to understand how these structures vary or agree in their concept-based input differentiation.

Altogether, in addition to the \textbf{conceptual method of using NAPs} for layer level neural network explainability, we implemented a \textbf{framework for extracting NAPs} from pretrained neural networks.
Furthermore, we designed an \textbf{interactive visualization environment for NAP analysis}.

\section{Related Work}

Most closely related to our approach are works in the field of neural network explainability, and more specifically, approaches for activation visualization.
In the following, we outline advancements to neural network understanding that these fields of research brought about and how they relate to the presented method.

\noindent\textbf{Neural Network Explainability.}
Neural network explainability can be defined as \emph{the ability to
provide explanations in understandable terms to a human}~\cite{doshi2017towards}.
Often, such explainability techniques make use of visualization, whereby Olah \ea{}~\cite{olah2018building} summarized the state-of-the-art techniques for visualizing the different components of a neural network.
Our work can be seen as an additional building block for visualizing learned layer concepts within neural networks.

One line of research in this direction is to visualize the location in the input data that exited individual units.
Some of the most prominent representatives of this class are saliency maps~\cite{simonyan2013deep} and their enhancements, such as Grad Cam~\cite{selvaraju2017grad} and Smoothgrad~\cite{smilkov2017smoothgrad}.
Saliency methods attribute neuron activations to locations in the input image by tracing a model's gradient.
Another approach to extract features is to modify the input in order to maximize the unit activation, so called feature visualization~\cite{erhan2009visualizing, olah2017feature, nguyen2016multifaceted, nguyen2016synthesizing}. 
While feature visualization can reveal interpretable results on a unit level, it is hard to interpret Feature Visualization on a layer level.

Concept-based explanation techniques have shown promising results in terms of understanding what neural networks learn on a layer level.
Yeh \ea{} \cite{yeh2021human} provide an overview of existing concept-based explanation techniques.
The importance of a concept for the prediction of a neural network can be tested with TCAV by training a linear model on the intermediate layer activations~\cite{kim2018interpretability}.
However, this manual process is geared towards testing, not discovery, and requires both gathering data for model training as well as an explicit query.
Although efforts have been made to automate concept discovery~\cite{ghorbani2019towards}, they are based on the assumption of segmented images with access to labeled data.
Furthermore, none of these techniques emphasizes the extraction of layer-level activation patterns as a direct explainability tool.
Rather, TCAV is an approach to test specific hypotheses with concept activation vectors, \eg{} \emph{how important is the stripes concepts for a zebra classification?}.
In contrast to TCAV where the focus is on testing for concept importance for specific queries, we provide a higher-level explainability approach in which layer concepts themselves are the focus of investigation.
Potentially, NAPs could even be used to discover concepts that can subsequently be tested for with TCAV.

\noindent\textbf{Activation Visualization.}
A subfield within the larger scope of neural network explainability is directed toward visualizing neural network activations.

Hohman \ea{}~\cite{hohman2019s} analyze network activations through aggregation.
They provide aggregation methods to discover important units for certain classes and compute the relation of neurons across layers.
Their focus is on using the highest activations resulting from the aggregations, while our approach applies pattern detection to find activation similarities independent on if the aggregated activation is high or low.
Furthermore, they analyze how units across different layers contribute to a predicted class output of the model.
NAPs, on the other hand, are not designed to explain classification predictions, but rather function as a high-level explainability tool for layer introspection.

To analyze layer level learned concepts, Bau \ea{} label each neuron of a layer with human interpretable concepts~\cite{bau2020understanding}.
Their method is to extract the highest activating sample per neuron and then label these samples with concepts.
This way, they extract what different concepts neurons of a layer have learned.
In contrast to our approach, they do not consider whether different neurons commonly fire together.
Similarly, NeuroCartography clusters neurons based on their spatial activation maps~\cite{park2021neurocartography}.
In turn, NeuroCartography can help to identify neurons that activate for similar features.
This can be used \eg{} to identify redundant neurons in the networks and support pruning approaches.
In contrast to our approach, it is not designed to surface layer level concepts.

Kahng \ea{}~\cite{kahng2017cti} presented a visual analytics tool to inspect the activation profiles of selected input examples alongside their associated classes.
Our work goes in an orthogonal direction, where activation patterns are computed automatically instead of showing activation patterns for manually selected samples.
We believe that our automated activation pattern discovery could augment existing capabilities in such tools.

\begin{figure*}[t]
\centering
\includegraphics[width=\textwidth]{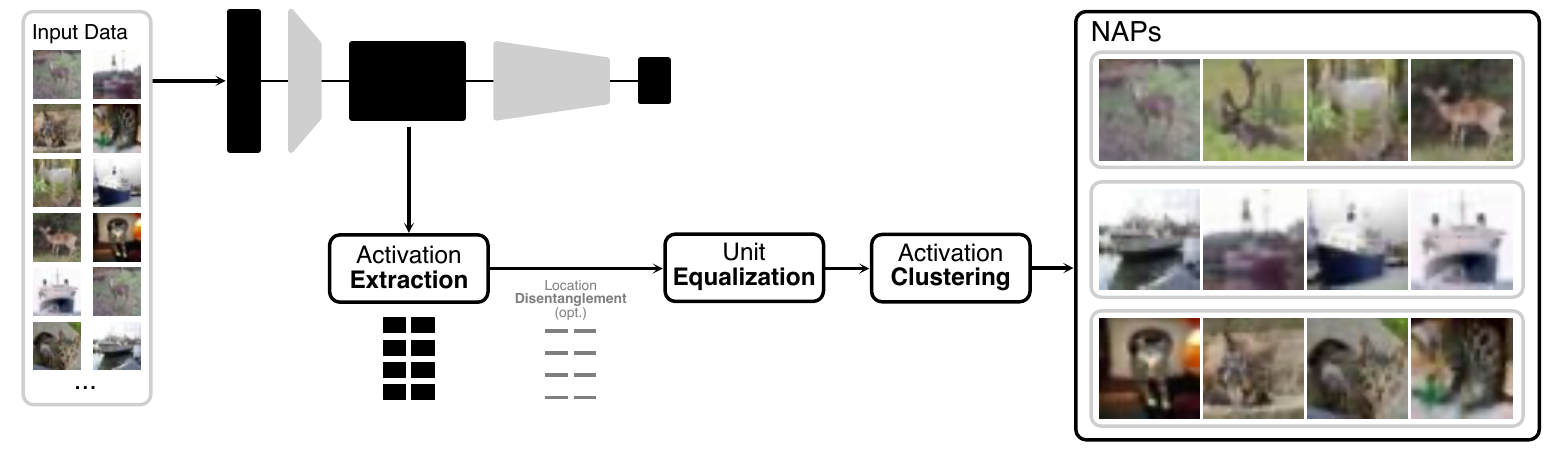}
\caption{
Overview of the pipeline for extracting neural activation patterns on a layer level for Cifar10.
First, one feeds input data to the model.
For each of these inputs, one extracts the activation values at the layer of interest (in this case, the second Conv2D layer).
Optionally, to disentangle NAPs from the activation location, these activation values can be aggregated to remove the spatial information.
Then, clustering is applied to obtain NAPs, which can represent different concepts the layer has learned.
}
\label{fig:NAP_overview}
\end{figure*}

To directly study semantically similar inputs based on neural network activations, dimensionality reduction techniques are often used.
Methods such as t-SNE~\cite{van2008visualizing} or UMAP~\cite{mcinnes2018umap} project points in the high-dimensional embedding space represented by the activations of a neural network layer down to two or three dimensions to make them inspectable.
Using these approaches, inputs that are close in the embedding space are mapped close together in the low-dimensional visualization space.
However, this removes a lot of information and requires manual clustering by the observer of the visualization.
Instead, we cluster the data directly on the basis of the high-dimensional layer activations.

\section{Neural Activation Patterns}

Each layer $l$ in a neural network produces an activation vector $a^l$ based on the input it processes.
Each unit of the activation vector can be thought of as describing a feature the layer tries to represent.
The combined activation of different units may represent an interpretable concept a layer has learned.
Interpretable concepts can be low-level, \eg{} \emph{stripes}, but such low-level concepts can also be combined to form higher-level concepts.
For example, the features \emph{animal}, \emph{legs}, and \emph{stripes} may form a \emph{zebra} concept.
Samples associated with a concept can be denoted as $x_i^{c}$.
The goal of our method is to find patterns in the neural activation of a model (NAPs) and to represent them in a way that allows them to be interpreted as concepts a layer has learned.
In the following, we describe how NAPs can be extracted from neural network models without any retraining or modifications through the steps outlined in \autoref{fig:NAP_overview}.

\subsection{Activation Extraction}
Given the layer of interest $l$ for network $n$ and input $x_i$, the output activation vector  $a^l_i$ of the sub network $n^l$ is extracted. 
For a dense layer, $a^l_i$ is a vector with one value per layer unit, while for a convolutional layer, each component of this vector represents a unit operating on the spatial domain of its input.
In turn, for the image case, components of $a^l_i$ also represent individual units but are two-dimensional matrices.
For a set of inputs $X$, we obtain a set of activation vectors for layer $l$:
 
\begin{equation}\label{eq:extraction}
    A^l = n^l(X)
\end{equation}

\noindent where $A^l$ can be seen as a high dimensional matrix with the transposed activation vector ${a^l_i}^T$ of input $x_i$ at row $i$.
There is no formal requirement on the input other than the ones set by the network itself.
It can be the training set, test set, subsets thereof, or completely separate data sets.

\begin{figure}[!b]
\centering
\includegraphics[width=\linewidth]{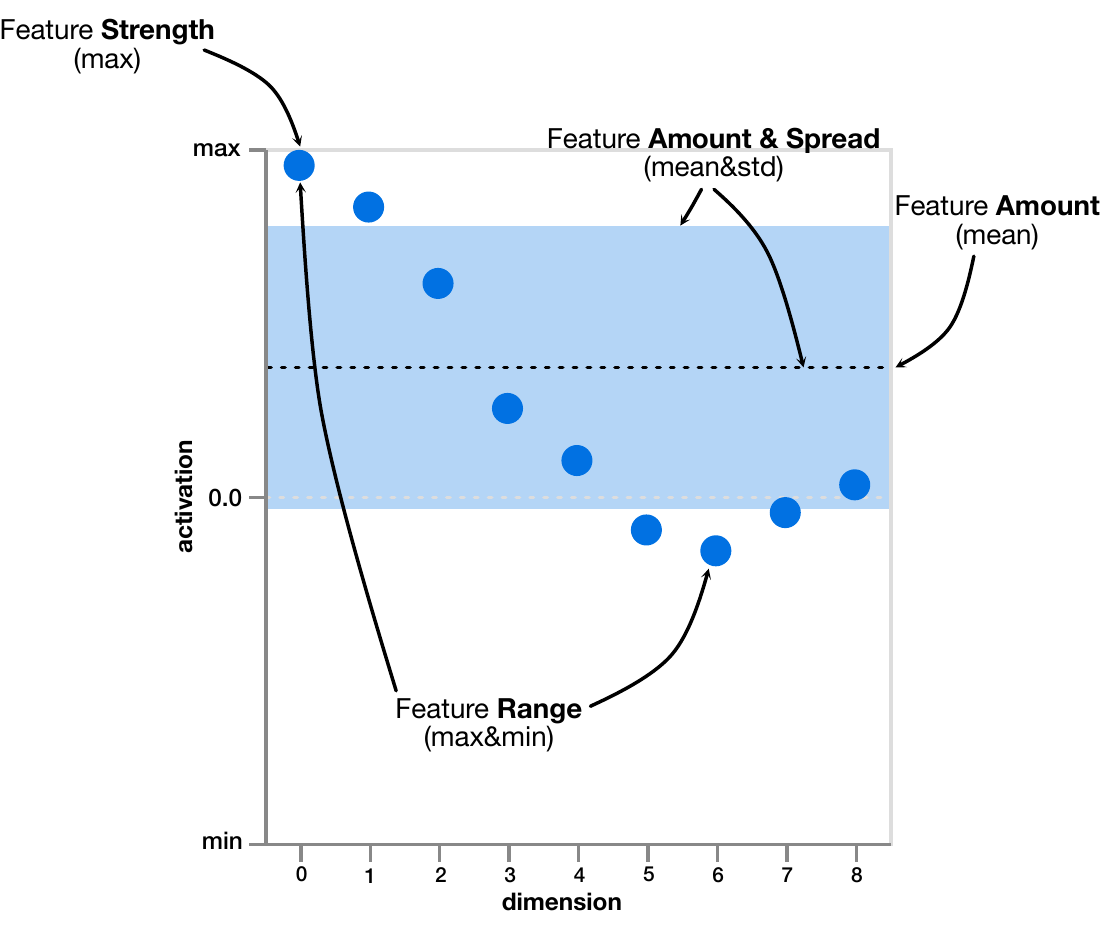}
\caption{
Illustration of the different methods for location disentanglement of a convolutional layer simplified to the one-dimensional case (would be two for images).
Horizontal axis depicts the spatial dimensions of a single unit within the activation vector.
Blue dots represent the activations of a single input for these spatial dimension. 
Feature strength: only the highest activation value is extracted.
Feature range: minimum and maximum activation value.
Feature amount: mean activation across all spatial dimensions.
Feature amount and spread: both mean and standard deviation.
}
\label{fig:aggregation}
\end{figure}

\subsection{Optional Location Disentanglement}

For the general case, extracting activations (\cf{} \autoref{eq:extraction}) is a sufficient basis for computing the NAPs of a model layer.
However, for convolutional layers, an activation vector $a^l$ does not only consist of one scalar value per layer unit.
Instead, for the case of image data, each element of $a^l$ is a two-dimensional matrix $M_{ij}$ of spatial activation values.
For the purpose of identifying concepts, the spatial information in the activation values can act as a confounding factor; animals are animals independent on if they are located in the bottom or top part of an image.
Thus, for convolutional layers, we remove the spatial information by aggregating the data across dimensions $i,j$, leaving only statistical descriptions of each unit within an activation vector unit $a^l_u$.

Different ways of aggregating will reflect different properties of the convolutional activation unit.
Extracting the highest activation value reflects peak feature strength.
Extracting both the lowest and highest activation values reflects the range in which the features are present. 
The mean activation instead reflects the amount or strength of the feature, e.g., appearing in many places increases the mean.
Adding standard deviation to the mean allows for differentiating between strong feature presence in a few locations and weak feature presence in many places.
These aggregation methods, all independent on the location of the feature, are detailed below and illustrated in \autoref{fig:aggregation}.

\begin{itemize}
    \item\textbf{Peak feature strength.}
        The maximum is extracted across all elements of the activation matrix, thus capturing the spatial location in the input that activated the unit the most.
        The input with highest activation generally tells us what the network considers most important.
        While this aggregation approach may sound promising, it can be compromised by relatively small, but strongly activating samples.
    \item\textbf{Feature range.}
        The minimum and maximum are extracted across all elements of the activation matrix.
        For layers with negative activations, this allows for taking opposite features into account.
        However, for layers with activation functions suppressing negative values the minimum value will in many cases be zero and therefore provide limited value.
    \item\textbf{Feature amount.}
        The average of all elements of the activation matrix is extracted, thus reflecting the amount of a feature.
        In contrast to max aggregation, patterns can also form for inputs with similar low activations. 
    \item\textbf{Feature amount and spread.}
        Uses the average in combination with standard deviation to disambiguate cases where otherwise a single high activation value would result in the same aggregated activation as many low activation values.
\end{itemize}

An added bonus of aggregating convolutional layers is that the computational cost decreases, which is beneficial for the method when it comes to its scalability with larger models.
We compare the described aggregation methods as well as NAPs computed without aggregation in \autoref{subsec:ablation}.

\begin{figure*}[ht]
\centering
\includegraphics[width=\linewidth]{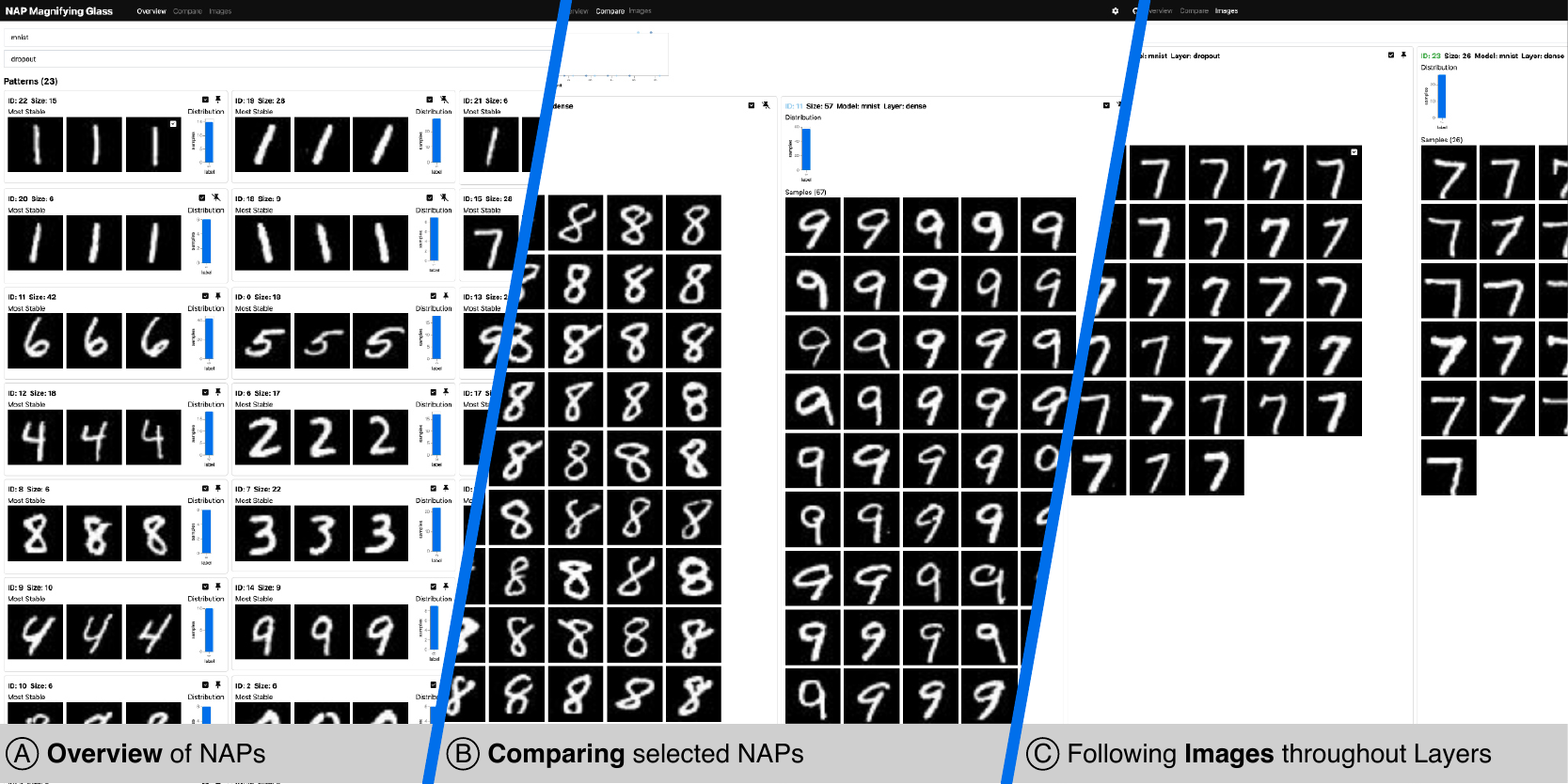}
\caption{
The NAP Magnifying Glass can be used to inspect and interpret NAPs.
It includes three main views.
A: Overview; NAPs for a selected model and layer are displayed through their most stable images and statistical metadata distributions (in this case only the label was selected for display).
B: Compare view; pinned NAPs can be inspected alongside activation distributions of the different patterns.
C: Image View; an image was selected to be able to follow its path through the network.
}
\label{fig:magnifying_glass}
\end{figure*}
\begin{figure}[!b]
  \centering
     \includegraphics[width=\linewidth]{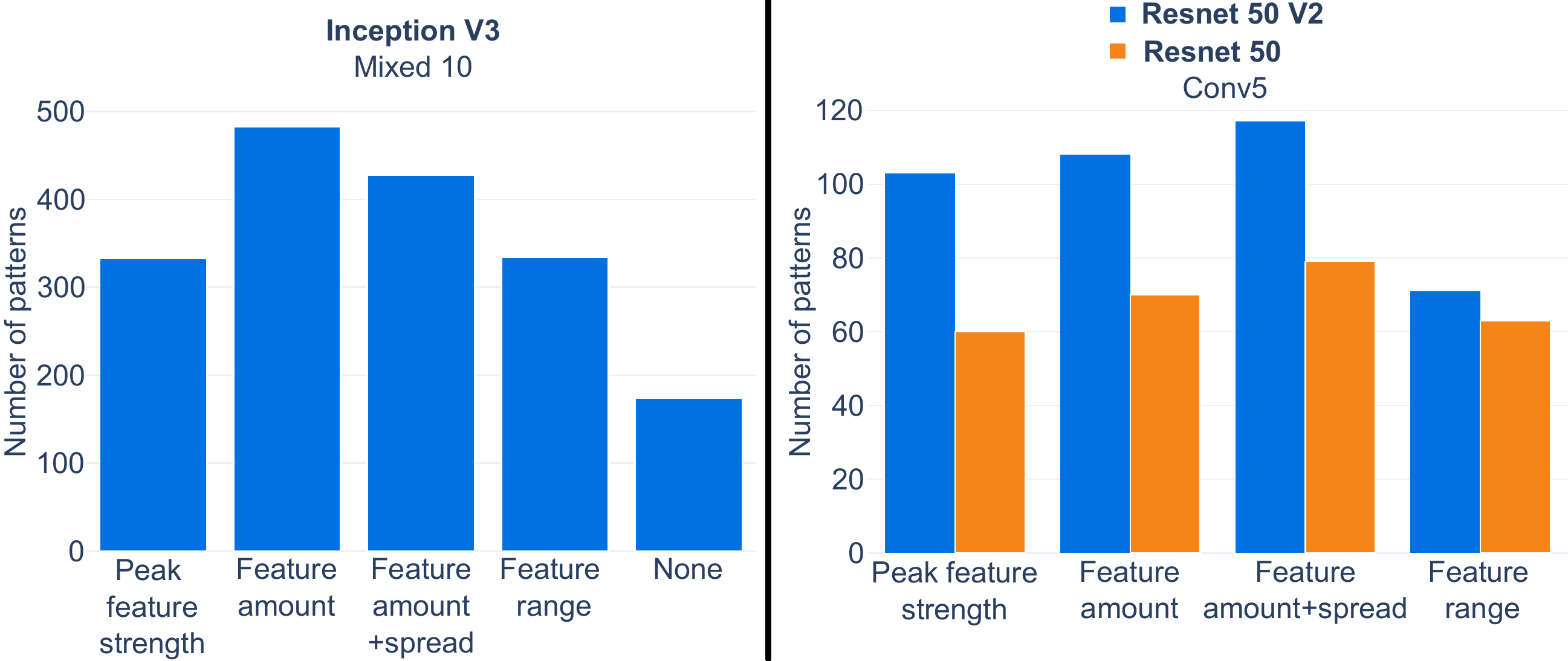}
  \caption{
  Number of patterns extracted for Inception V3 (left), Resnet 50, and Resnet 50 V2 (right) for different methods of location disentanglement. 
  Few patterns are extracted when spatial information is kept (None).
  }
  \label{fig:inception_resnet50_resnet50v2_aggregation}
\end{figure}

\subsection{Activation Unit Equalization}
Naturally, each unit of the activation vector does not necessarily work on the same scale as the other units in the activation vector. 
For example, one unit may produce activations on the domain $[0~1]$, while another one does it in the $[0~10]$ domain. 
Subsequent layers may change these scales through their weights, and since a unit is connected with multiple weights to the following layer, there is no clear weighting of the unit activation either.
In turn, a highest value of 1 in one unit may be equally important to the highest value of 10 in another unit for subsequent layers.
As a result, the importance of the relative scales between units in a layer is unknown.
When extracting concepts based on unit activations for a specific layer, it is thus necessary to normalize the domains of the different units to remove any bias towards units with larger activation ranges. 

For this normalization, we need to consider the context in which it is performed.
Unit activations can be positive, representing the presence of a feature (\eg{} \emph{straight line}) and negative, representing a different, often opposite feature (\eg{} \emph{curved line}).
On the other hand, a zero activation value represents the total absence of the feature (\eg{} \emph{no line}).
To preserve this inherent meaning, the normalization must not shift activation values away from zero while scaling positive and negative activations by the same amount.
Therefore, we normalize each unit by its maximum absolute activation across all data samples $X$:

\begin{equation}\label{eq:normalization}
\hat{a}^l_{i,u} = \frac{n^l_u(x_i)}{max(abs(n^l_u(X)))}
\end{equation}

\noindent where $n^l_u$ represents the activation of one unit $u$ extracted from network layer $l$.
This way, units that can only activate positively are normalized between $0$ and $1$, whereas units containing negative activations are normalized between $-1$ and $1$.
In the following, we will always assume activations are normalized according to~\autoref{eq:normalization}.

\subsection{Activation Clustering}
The underpinning idea behind NAPs is that inputs that share the same features activate layer units in a similar way, and thus form an activation pattern.
Therefore, based on the normalized activations $A^l$, our objective is to find the sets of inputs $x_i \in X$ that produce similar activations.
As each input is represented by an activation vector in $A^l$, this problem can be formalized as finding subspaces that uniquely describe a set of inputs. 
Finding patterns in high-dimensional data is a well-studied problem with a range of solutions~\cite{parsons2004subspace,xu2005survey,xu2015comprehensive}.
Our method does not put a restriction on the pattern extraction algorithm.
However, important properties for the pattern extraction algorithm are, that it 1) requires few or preferably no parameters, 2) that it is computationally efficient and scales to the high dimensions spanned by the activation vector, and 3) that it can deal with noise. 
In addition, algorithms that provide guidance on the relative strength of discovered patterns are can aid in pattern exploration.
For our experiments, we employ HDBSCAN~\cite{McInnes2017} as a clustering algorithm incorporating these properties, see \autoref{sec:implementation} for details. 

After activations have been extracted, disentangled from location information, and normalized according to \autoref{eq:normalization}, activation subspaces are found using clustering algorithm $s$.
Since activations are normalized on a per unit-basis, we can employ a standard Euclidian distance metric for computing activation vector differences.
By clustering $A^l$, the sets of inputs $x_i \in X$ producing similar activations are grouped together.
Based on the assumption that inputs that share the same features activate layer units in a similar way, we assume that individual data samples that are clustered together can be used as representatives of such feature combinations, which, in turn, can be interpreted as a concept $c$:

\begin{eqnarray}\label{eq:naps}
    s(\hat{A}^l) \rightarrow NAP^{c_j} \approx  \{x_i \in X | x_i^{c}\} 
\end{eqnarray}

\noindent where $x_i^c$ represents an image that is a representative of concept c.
Optimally, there would be no other NAP capturing the same concept.
For example, if there is a \emph{straight} line concept that a layer has learned, a perfect NAP for that layer would include all inputs that contain the concept (\ie{} $NAP^{straight} = \{x_i \in X | x_i^{straight}\}$).
Furthermore, the perfect NAP would not contain additional inputs and there would be no second NAP depicting the same concept.
However, in practice, obtaining perfect NAPs would imply using input data perfectly reflecting a concept as well as a perfect clustering algorithm.
Because neither real-world data that perfectly reflects a single concept, nor a perfect clustering for concepts exist, perfect NAPs cannot be obtained.
Nevertheless, in our practical experience, the presented automatic NAP extraction produces interpretable concepts and can, thus, serve as a building block for human-in-the-loop exploration to understand what a neural network layer has learned.

The representation of a NAP follows from its definition (\autoref{eq:naps}).
It includes a collection of input examples that are associated with a concept $x_i \in X | x_i^c$ as well as the activation distribution $A^l$ of samples that belong to a NAP.
As such, NAPs can be visually analyzed as explained in \autoref{sec:nap_magnifying_glass}.

\section{Implementation}\label{sec:implementation}
We implemented both a method for extracting NAPs and a visualization interface for inspecting NAPs.

\subsection{NAP Extraction}
The presented method has been implemented as a Python package using the TensorFlow framework to extract NAPs from pre-trained neural network models.
This Python package allows users to obtain NAPs simply by providing a trained neural network model, supplying sample data, and specifying for which layers of the model to extract NAPs.
In the following, we describe how this package was implemented to work in unison with human intuition on explainability.

\noindent\textbf{Model and data processing.}
A user can obtain NAPs for their own model and dataset without any configuration as follows:

\begin{lstlisting}[language=Python]
import nap

naps = nap.NeuralActivationPattern(ml_model)
naps.extract_patterns(input_data, layer_name)
naps.magnifying_glass(destination)
\end{lstlisting}

The code above sets up the method for extracting NAPs using the provided neural network model, extracts NAPs for the provided layer, and then exports them to be visualized and examined in the magnifying glass exploration tool.
Parameters, such as the method to be used for location disentanglement, can optionally be set upon NAP initialization.

\noindent\textbf{Clustering Algorithm.}
The most critical step in the process of NAP extraction is pattern detection.
More specifically, the choice of the clustering algorithm and its settings plays an important role in the way NAPs are extracted from a model.
Based on the three criteria for pattern extraction, i.e., few parameters, computational efficiency, and robustness to noise, we investigated several options for the clustering algoritm to be used.
As the number of concepts in a layer is unknown, all algorithms that require the specification of a fixed number of clusters, such as K-Means, were disregarded. 
When it comes to performance DBSCAN scales reasonably well, but requires fine-tuning the density parameter, which varies between different models and data sets.
Hierarchical density-based spatial clustering (HDBSCAN)~\cite{McInnes2017}, on the other hand, automates the parameter tuning while also outperforming DBSCAN in terms of performance.
Additionally, in our experiments it seemed to be able to deal with a variety of models and input data with the default minimum cluster size value of five.
To make our approach independent of the data set at hand, we, thus, chose to use HDBSCAN in our implementation.
To obtain smaller and more homogeneous clusters, we choose the leaves of the cluster tree instead of using the default excess of mass method. 
As a result, instead of capturing coarse concepts that are hard to understand from visual inspection, we capture more fine-grained concepts.
As the amount of layer activations generated for large models and data sets can require significant memory, activation values can optionally be cached and read from disk using HDF5\footnote{//https://www.h5py.org/}.
In \autoref{subsec:ablation}, we describe how we evaluated different parameter settings for the clustering we employ.

\noindent\textbf{NAP Content.}
After activations for a layer have been clustered, we extract NAPs based on these clustered activations.
As described in \autoref{eq:naps}, a NAP consists of a set of data points that produce similar activation profiles.
In turn, this set of input examples can be used to represent a NAP.
For detailed inspection , we also provide unit-level statistics about the activation profile of each NAP; this includes the average activation value, activation quartiles, and the lower as well as upper bound of the data input to the clustering algorithm on a unit basis.
When there is additional information for individual data points, \eg{} labels, or model predictions, these might also be taken into account when inspecting and interpreting a NAP.
Note that such additional information is not used for computing NAPs, which means that the implementation works for many neural network architectures and tasks, not only for labeled image data.

\subsection{NAP Magnifying Glass}\label{sec:nap_magnifying_glass}
Alongside the method for extracting NAPs from pretrained neural network models, we also provide a visualization interface for inspecting NAPs.
This interactive visual environment is implemented as a web-based visualization interface.
The frontend for the NAP Magnifying Glass is implemented in Svelte\footnote{https://svelte.dev}, whereas the backend providing data access is based on Python and Flask\footnote{https://flask.palletsprojects.com/}.
The NAP Magnifying Glass, which can be seen in~\autoref{fig:magnifying_glass} includes different views for analyzing NAPs.

\noindent\textbf{Layer Overview.}
For the main goal of inspecting layers of neural networks, the layer overview shows NAPs for a selected model layer as can be seen in \autoref{fig:magnifying_glass} A.
Here, NAPs extracted for a layer are shown through representative images and activation statistics.
To guide the human interpreter to the most stable NAPs first, NAPs are sorted according to cluster persistence.
Similarily, within a NAP, each data sample $x_i$ is ordered according to its persistence within the cluster.
Thus, it becomes possible to inspect which NAPs and inputs are most likely to represent a concept. 
NAPs can be filtered using their metadata, such as model predictions, and, in case of supervised learning, labels.

\noindent\textbf{Compare View.}
NAPs can be selected across different models and layers for more detailed inspection and comparison, as shown in~\autoref{fig:magnifying_glass} B.
Such selected NAPs are displayed side by side, whereas all samples that belong to the NAP under investigation are visualized in this view.
In addition to the input images represented by each NAP, the view also shows their unit activation profiles.
For the last layer, where each activation unit corresponds to a label, the statistics can be used to inspect which types of inputs may lead to similar predictions.

\noindent\textbf{Image View.}
The third view in the NAP Magnifying glass, depicted in~\autoref{fig:magnifying_glass} C, serves as a means to follow images through different layers of a model.
When selecting an image, all NAPs of the model under investigation that contain the selected image are displayed.
Hereby, we sort layers so that early model layers are displayed to the left, whereas later layers are displayed to the right.
This way, users can investigate how the models understanding of an image type is built throughout successive layers.

Throughout the visualization interface, incorrect predictions are highlighted with an icon to facilitate analysis of model prediction.
This enables the investigation of concepts that may be confusing to the model.

\section{Results}\label{sec:results}

Using the implementation presented in~\autoref{sec:implementation}, we conducted several parameter studies evaluating the approach with different settings.
Following these results, we will also present qualitative results we obtained with our method using different neural network models.
All of these experiments were conducted with different models, two smaller models we trained ourselves on MNIST~\cite{lecun2010mnist} and Cifar10~\cite{Krizhevsky09learningmultiple}, respectively, and three models pre-trained on ImageNet~\cite{deng2009imagenet}, namely Resnet50~\cite{he2016deep}, Resnet50 V2~\cite{he2016deep}, and Inception V3~\cite{szegedy2016rethinking}.
To test our approach, we generated NAPs on the basis of subsets of the original training, test, or validation data. 
Unless otherwise specified, feature amount location disentaglement with 1\% of the 1.2M training images were used for the ImageNet-based models, while the complete test set was used for CIFAR10 and MNIST.
The computation time for generating NAPs according to \autoref{fig:NAP_overview} was in the order of minutes for the small models and about an hour for the large models per layer.

\begin{figure}[t]
  \centering
     \includegraphics[width=\linewidth]{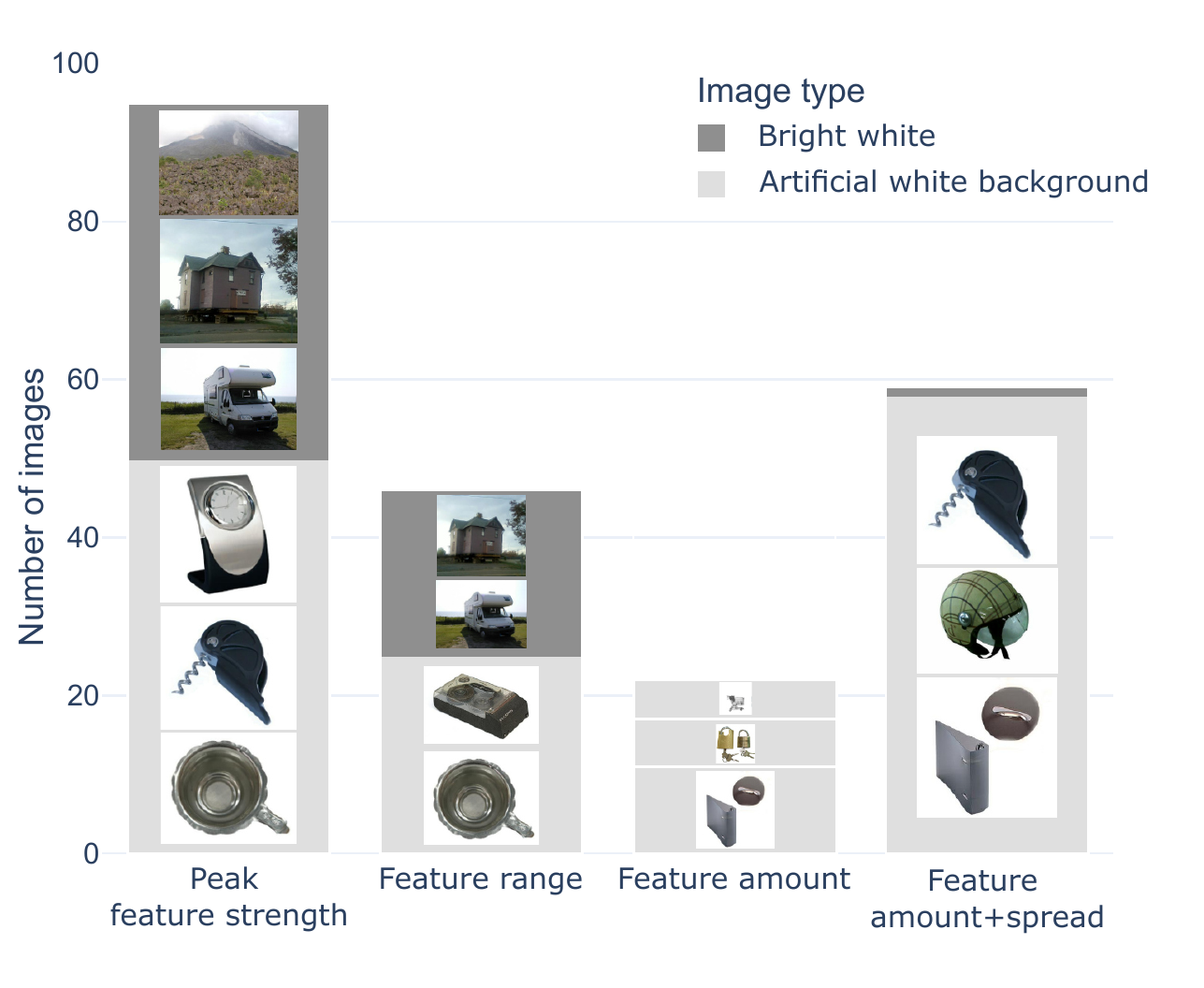}
     \caption{
     Number of images in NAPs including \emph{artificial white background} for the \emph{mixed 1} layer in Inception V3 using different location disentaglement methods. 
     Images inside the bars are examples from the respective NAP.
     The peak feature strength and feature range methods result in a single pattern including images with \emph{artificial white background} and \emph{bright areas}. 
     The feature amount location disentanglement method results in three different NAPs with slight variations of images, while the feature amount and spread method captures most images with \emph{artificial white background} within the same NAP.}
  \label{fig:aggregation_white_background}
\end{figure}

\begin{figure}[!b]
\centering
\includegraphics[width=\linewidth]{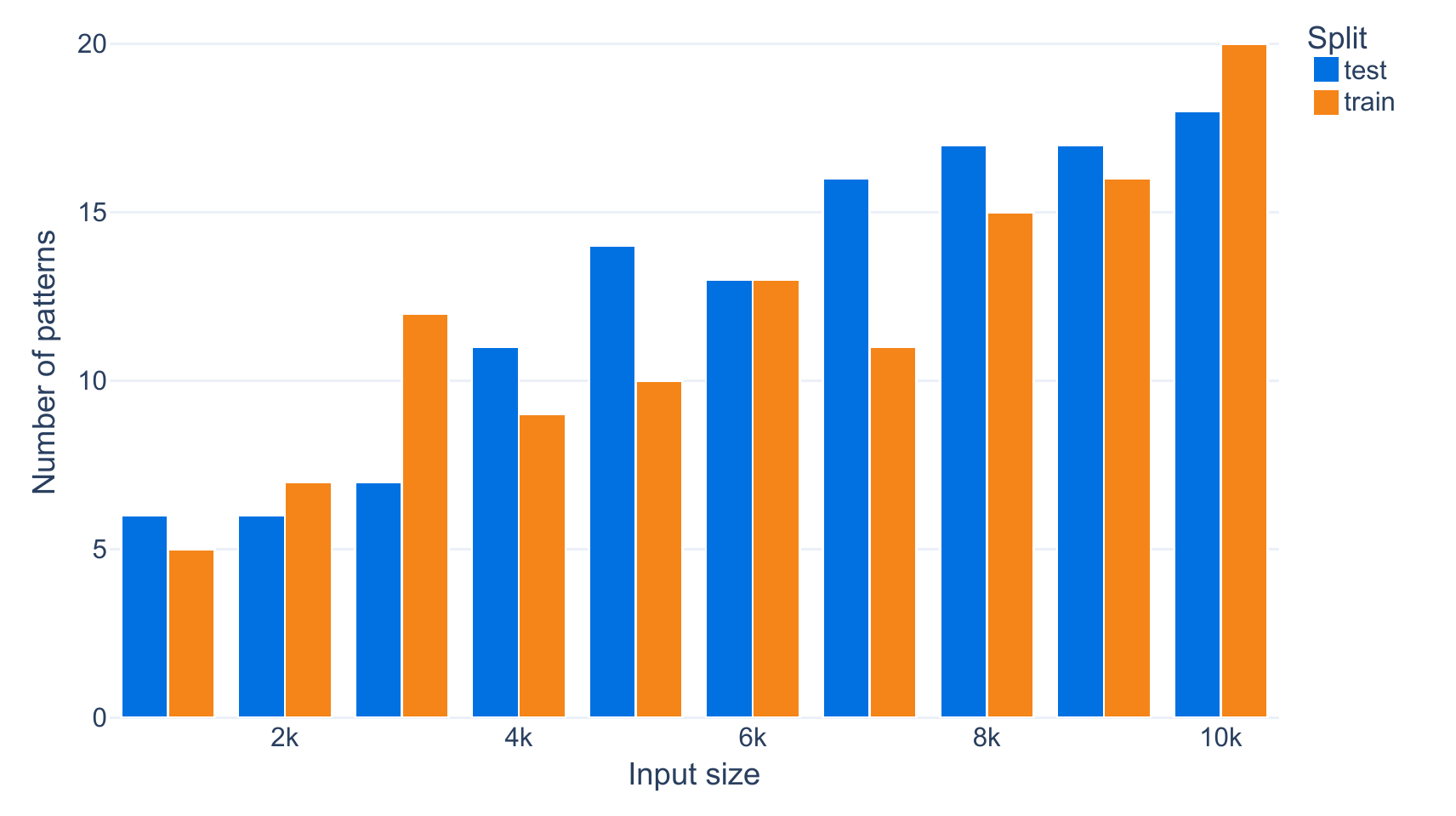}
\caption{
Number of extracted NAPs for increasing number of inputs for the \emph{dense} layer of the CIFAR10 model.
The number of extracted NAPs generally increases with increasing input sizes. 
The balanced train and test sets follow the same overall trend. 
}
\label{fig:cifar10_dense_input_size}
\end{figure}

\subsection{Parameter Studies}\label{subsec:ablation}
The two most important parameters of the presented method are the choice of location disentaglement method and the input data itself.
Besides these two, we also evaluated the main clustering algorithm parameters' \emph{minimum cluster size} and \emph{minimum samples}.
These two parameters control the smallest possible number of similar samples and how dense samples need to be packed in order to be considered a cluster.
Here, we found that leaving the default parameters set to five worked well for the tested models and input sizes. 
If fewer, but larger patterns are desired, it is possible to tune the clustering parameters, but the default parameters have been used for all the examples in this work.

\begin{figure*}[t]
  \centering
     \includegraphics[width=\linewidth]{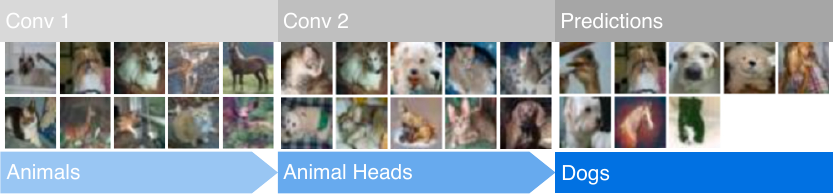}
  \caption{
  Following the dogs prediction images through the network for Cifar10.
  NAPs evolving from brown and white animals over furry animals to the selected dogs NAP in the prediction layer.
  }
  \label{fig:dogs_cifar}
\end{figure*}

\begin{figure*}[t]
  \centering
     \includegraphics[width=1\linewidth]{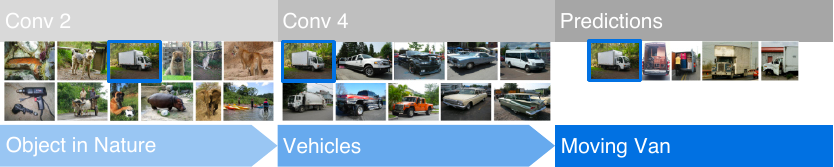}
  \caption{
  Following a single image of a moving van through the layers of Resnet50.
  }
  \label{fig:moving_resnet}
\end{figure*}

\noindent\textbf{Location Disentaglement.}
The number of NAPs extracted by each location disentaglement method for the ImageNet validation set is depicted in \autoref{fig:inception_resnet50_resnet50v2_aggregation} (left).
It can be seen that using \textit{feature amount} or \textit{feature amount and spread} leads to the most extracted NAPs. 
This trend also applies to other layers, please confer the supplementary material for detailed results.
Interestingly, comparing the two ResNet models in \autoref{fig:inception_resnet50_resnet50v2_aggregation} (right) shows that more NAPs are found in the higher accuracy ResNet50 V2 model even though both models contain the same amount of weights. 
We speculate that this is either due to architectural changes or because the model has learned more.
As anticipated, without location disentaglement, indicated by \emph{None} in \autoref{fig:inception_resnet50_resnet50v2_aggregation} (left), fewer NAPs are extracted. 
Furthermore, NAPs extracted without location disentaglement are, in most cases, extracted with the other methods as well.
In general, the most stable NAPs tend to be found using any of the methods.

When inspecting NAPs in detail, the \textit{peak feature strength} location disentanglement method tends to generate less well separated NAPs that include larger numbers of images compared to the other methods. 
For example, for the early \emph{mixed 1} layer in Inception V3, images with artificial white background are separated into their own NAP(s) for \textit{feature amount} and \textit{feature amount and spread}, while using the \textit{peak feature strength} method results in a pattern containing images with both white artificial background and images with white regions, as illustrated in \autoref{fig:aggregation_white_background}.
The \textit{feature range} method also has the trait of merging such concepts and also produces smaller NAPs.
Out of the evaluated methods, the \textit{feature amount and spread} best separates the artificial white image background images from other images that contain white content.

\noindent\textbf{Input Data.}
The number of extracted NAPs for varying amounts of input data and different data set splits for the dense layer of the CIFAR10 model can be seen in \autoref{fig:cifar10_dense_input_size}. 
As both the training set and test set are equally balanced we expect the number of extracted NAPs to be roughly the same. 
We visually verified that the extracted NAPs are similar, e.g., that red trucks are extracted in the second convolutional layer for both the train and test sets.  
Of course, there are also differences, such as a night image pattern extracted from in the training set, but not from the test set.

With larger amounts of input data, more fine grained NAPs are generally extracted.
Additionally, most layers follow the trend of increasing numbers of NAPs for increasing amounts of input data.
However, some layer types, such as the flatten layer do not follow this trend.
The reason for the flatten layer going against this trend is that it simply reshapes the convolutional layer into a one dimensional vector, and we do not perform spatial disentanglement on such vectors.
As described in the previous paragraph, no location disentanglement generally reduces the number of extracted NAPs.
Altogether, we observed similar results for the larger networks, for details, confer to the supplementary material.

\subsection{Qualitative Results}
Following these parameter studies, we also used NAPs as a  qualitative exploration tool.
Starting with small, self-trained neural networks, we will work our way up to the well-known Inception network, whereby we will present insightful NAPs for each of the networks we investigated.
Finally, we will compare a set of NAPs between three models trained on the ImageNet dataset, namely Resnet50, Resnet50 V2, and Inception V3.
First, we provide the results of a small informal user study that serves as a basis for the naming of all concepts in the paper.

\begin{figure}[b]
  \centering
     \includegraphics[width=\linewidth]{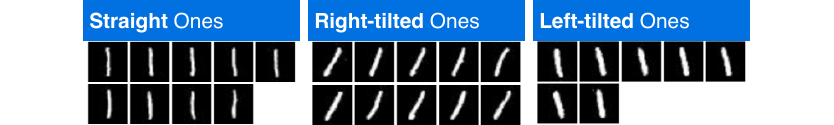}
  \caption{
  This first convolutional layer of the MNIST model already differentiates between \emph{straight}, \emph{right-tilt}, and \emph{left-tilt} lines.
  }
  \label{fig:ones_mnist}
\end{figure}

\noindent\textbf{Concept Perception.}
To test whether the extracted concepts were perceived similarly, we conducted a small-scale, informal user study with 8 ML practitioners.
We sent the NAPs shown in \autoref{fig:teaser}, \autoref{fig:ones_mnist}, \autoref{fig:dogs_cifar}, and \autoref{fig:moving_resnet} to participants after removing the layer and concept labels.
Then, we asked participants to assign a short title to each of the image groups.
Overall, participant agreement was high, \eg{} for the leftmost NAP in \autoref{fig:ones_mnist}, answers were either along \emph{straight line} or \emph{straight one}, and for the NAPs in \autoref{fig:teaser}, most participants labeled them as \emph{animals that can fly} or \emph{winged animals}, followed by \emph{butterflies}, and \emph{yellow butterfly}.
For full results of this study, we refer to the supplementary material.
We assigned concept names for all these figures based on the feedback from participants.
An exception to this are the respective prediction layers, where we use the actual model prediction as a label for the NAP.

\noindent\textbf{Smaller Networks.}
We trained a neural network with two convolution layers each followed by max-pooling and one fully connected layer for the final predictions on the MNIST data set.
For this model, we found that early layers already disambiguate different ways of writing numbers.
Looking at \autoref{fig:ones_mnist}, one can see that in the first layer already, the model picks up different concepts of writing the number one.
In this case, these NAPs are differentiated by the tilt of the written numbers.
One can see, that these NAPs represent three distinct tilt angles, one that tilts numbers to the left, one that tilts numbers to the right, and one for straight numbers.

To understand how the Cifar10 model (three convolutional layers, two fully connected layers) arrives at its classification for dogs, we find NAPs with dogs in the different layers (\autoref{fig:dogs_cifar}).
We see that in the first convolutional layer of the network, many of these dog images are included in a NAP that contains \emph{animals}.
The second convolution layer seems to refine this, separating these images and providing a NAP with only \emph{animal heads}.
In the prediction layer, all shown images are classified as \emph{dog}.
However, there is one misclassified bird and one misclassified horse among the dogs, which means that the model or training data need to be adjusted to be able to differentiate these concepts.

\noindent\textbf{Resnet50.}
Here, we show how different layers of the model operate on separate kinds of concepts.
To illustrate this, we follow the moving van shown in~\autoref{fig:moving_resnet} across layers.
We can see that it is first associated with a NAP of \emph{objects in nature}, before being narrowed down to a \emph{vehicles} NAP.
In the prediction layer, the image ends up in the \emph{moving van} NAP with other correctly classsified moving vans.

\begin{figure}[t]
  \centering
     \includegraphics[width=\linewidth]{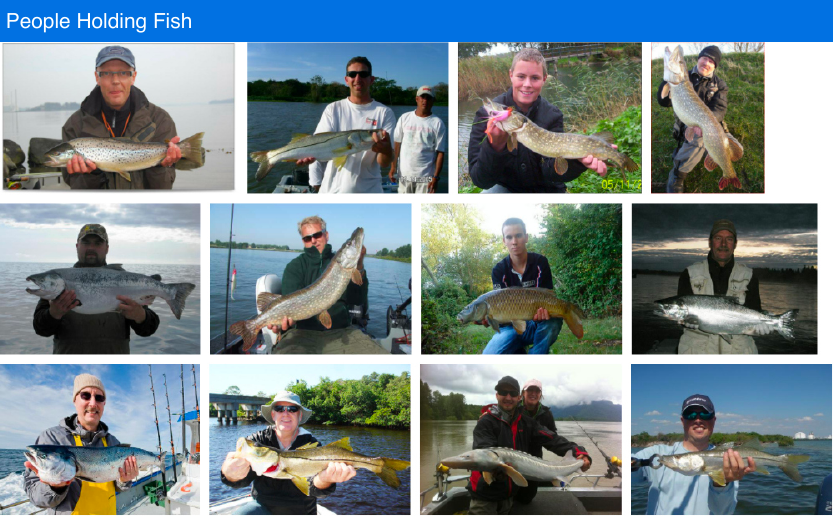}
  \caption{
  Through NAPs depicting people holding fish, we found that this is a common concept in ImageNet.
  }
  \label{fig:people_fish}
\end{figure}

\noindent\textbf{Resnet50 V2.} 
Here, we demonstrate how to interpret NAPs to gain insights into model predictions and how they could relate to data collection.
In \autoref{fig:people_fish}, we show a NAP found in the conv5 block of Resnet50 V2, which is a a late layer in the architecture.
The images depict \emph{people holding fish}, and we found that some NAPs in the prediction layer for the fishes also only contain people holding fish rather than fish under water.
This could be problematic as the model might not have learnt the concept of fish, but instead the context around it. 
Inspecting the ImageNet training data for tench fish, we can indeed see that almost all images are of people holding the fish.
This phenomenon has been reported by other researchers as well~\cite{hohman2019s,tench}, confirming the validity of our findings using NAPs.
Thus, we can see that NAPs can provide insight into the way data was collected, leading to insights on overrepresentations of data.
Similarly, concepts that are not represented by NAPs, but would be expected to exist, might indicate that data for a specific concept is missing, which can also be an indicator for problematic bias.

\begin{figure}[b]
  \centering
     \includegraphics[width=\linewidth]{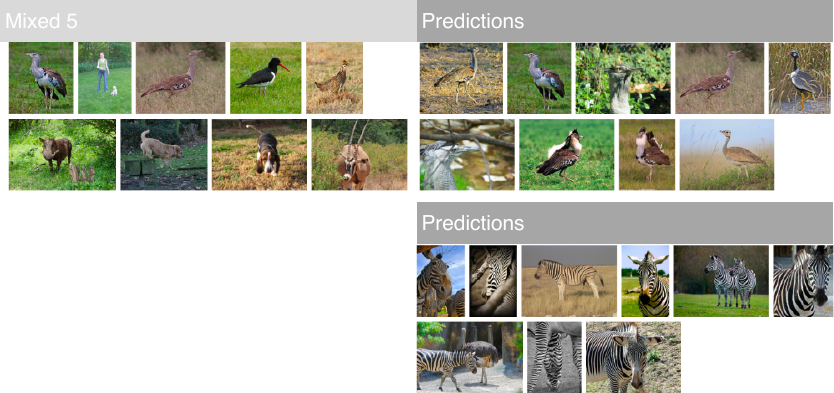}
  \caption{
    Some images appear both in NAPs of early layers and late layers, while oter NAPs only include images that never appear in early layers of a model
  }
  \label{fig:inception_concepts}
\end{figure}

\noindent\textbf{Inception V3.}
In~\autoref{fig:inception_concepts}, one can see how some images are included in NAPs beginning in early layers, while other images only occur in NAPs of late layers.
The reason for this might be that some concepts need to be assembled from important concepts that span a variety of inputs early on, while other concepts can only assemble late in the network since the images that they contain do not share important low-level features.

\begin{figure*}[h]
  \centering
     \includegraphics[width=\linewidth]{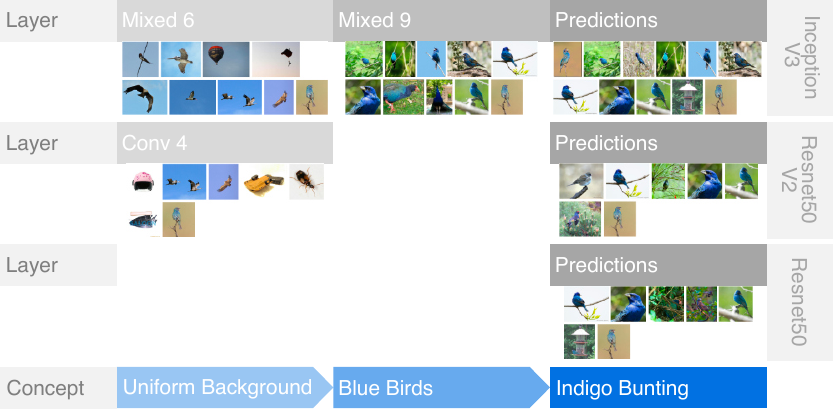}
  \caption{
  Following images predicted as indigo bunting through different models, namely Resnet50, Resnet50 V2, and InceptionV3.
  }
  \label{fig:bunting_compare}
\end{figure*}

\noindent\textbf{Comparing Models.}
NAPs can also be used to compare different models by analyzing the concepts discovered across layers.
In~\autoref{fig:bunting_compare}, we follow images classified as indigo bunting for three models, namely Inception V3, Resnet V2, and Resnet.
One can see how, the more advanced the model, the more the classification is built up from earlier layers.
The inception network contains NAPs that include images of indigo buntings for many layers.
Building up from uniform background images over blue birds, to the indigo bunting class.
For Resnet50 V2, only two layers include images of indigo bunting, whereas we could not find the blue bird concept in this network.
Resnet50 even misses the solid backgound NAP with the indigo bunting.
This might correlate with the fact, that all indigo buntings in the Inception V3 prediction layer NAP got classified right, while the NAPs for the final layer of Resnet contain images classified as indigo bunting that are in fact other bird species.

\section{Discussion}

NAPs can serve as a layer-level explainability method that can be used to inspect how neural network layers react to different kinds of inputs.
As such, NAPs provide an additional building block in the line of neural network explainability work.
While there exist a myriad of methods for investigating individual units, the concepts learned by entire neural network layers have not seen as much attention with regard to visual introspection methods.
As shown in \autoref{sec:results}, NAPs, which serve as a layer-level explainability tool are a promising candidate to fill this gap.

While NAPs can be exported for a neural network layer automatically, they still require human interpretation.
Our qualitative small-scale study showed that ML practitioners tend to agree on the concepts that are represented by NAPs.
However, similar to other explainability techniques, even if human annotators agree on a concept, it might not always reflect what the model sees.
Thus, we argue for combining different explainability techniques, such as using NAPs for testing with TCAV, or combining attribution techniques with NAPs.

Discovering NAPs also depends on the input data that is used to investigate a model layer.
If the input data does not include a concept, it will not be possible to extract a NAP for it, even if the model has learned said concept.
Furthermore, as we are using a density-based clustering algorithm, different data subsets can lead to variable NAPs.
As our aim is to provide a general explainability method that can be used for qualitative analysis, we do not think that this harms the utility of NAPs.
In fact, ML practitioners can craft data sets specifically to discover NAPs on a different level, such as using all classes vs. just providing input examples of one class.

When it comes to selecting parameters for generating NAPs, we found that the default clustering parameters work well.
Thus, we have used them to generate the results throughout this paper. 
Out of the four spatial disentaglement methods used for convolutional layers, it was shown that \textit{feature amount} is most suitable for extracting many NAPs, while the \textit{feature amount and spread} resulted in the most discriminative NAPs.

We found that earlier layers tend to generally produce less NAPs that are also harder to interpret.
The reason for this drawback might be that earlier layers are considered to represent more abstract, low-level features, which would naturally cover a broader spectrum of examples.
In later layers, it might be possible to represent image types by few high-level features (\eg{} \emph{Grass}, \emph{fur}, and \emph{dog snout}).
However, similarly narrow representations might not be easily obtained for lower-level features.
If \eg{} we assume a layer responds strongly to the concepts of colors and line segments, it is unlikely that there are groups of images sharing these low-level features in the same way (\ie{} these dog images might contain very different line segments as well as fur, sky, and background colors).
If the focus is more on low-level concepts, it might be neccessary to include images that contain \eg{} uniform colors and simple patterns, which might reveal more early-layer but less late-layer NAPs.

NAPs are not designed to replace other explainability tools, but rather as an augmentation of them.
Since NAPs, in contrast to other methods, are designed to work on a layer-level, we hope that they can provide insights that other explainability methods could not.
We think the way explainability methods are often simply applied on a layer-level, especially max activations and Feature Visualization, might need to be overthought.
As such methods cannot surface the complex interdependence of features within a layer, we advise using other methods, such as NAPs or TCAV, instead (see~\autoref{fig:ActivationDistribution}).

Altogether, NAPs in isolation can neither provide a complete set of concepts the model has learned, nor can they be used for quantitative evaluation of a model.
However, similar to other explainability tools, \eg{} Feature Visualization, NAPs can provide a unique insight into neural network layers.
In turn, our understanding of these models can be further deepened using such visual introspection techniques.

\section{Future Work}
While NAPs are a first step towards identifying different concepts that a neural network layer has learned, there are numerous ways forward with NAPs and other layer-level explainability methods that could provide further insights.

\noindent\textbf{Slicing Directions.}
If activations are sliced across different dimensions, one can answer separate questions about the model and data.
For example, if one slices across classes, one may find intra-class differences, similar to Wei \ea{}\cite{wei2015understanding}.
Another slicing direction of interest could be investigating unit-level NAPs which would be closer to existing work on unit-level explainability.

\noindent\textbf{Data Types and Tasks.}
While we tested our approach on image data, it could be applied to different data types since clustering activations is universally applicable.
However, one would need to modify our presented visualization interface to reflect different data types.
Similarly, extracting NAPs is also not limited to classification tasks.
NAPs could be extracted for any type of neural network, such as GANs, Transformers, and RNNs, as long as it contains activation vectors for individual layers.

\noindent\textbf{NAPs with Attribution.}
One drawback of our current approach of extracting NAPs is that is it up to human interpretation to resolve which features the samples that represent a NAP share.
If NAPs were combined with attribution methods, it might be possible to at least guide the human to the structures in the input data that are most representatinve of the NAP.
However, as NAPs are not based only on high activations, current attribution methods that are based on activation maximization would need to first be modified to work with NAPs.

\section{Conclusion}
In this work, we present NAPs, a new method for extracting learned concepts from neural network layers.
The method works by first feeding a set of inputs to a pre-trained model and capturing the activation output of a layer.
These activations are then normalized on a per-unit basis to equalize the importance of each unit before being sent to a clustering algorithm that extracts NAPs.
Extracted NAPs are associated with their corresponding inputs and according metadata.
NAPs can thereby be used to understand how different layers of a neural network model process data.
We showed how inspecting NAPs within a layer and across layers could lead to many different types of insights related to the ML model.
For example, in a quantitative analysis, we investigate how NAPs could be used to discover insufficient model performance or bias in the training data.
A particularly promising feature of the NAPs is that they allow for model comparison even for models with different architectures.
For broad adoption and reproducibility, we provide a package for extracting NAPs alongside a visualization interface that facilitates their analysis.
In conclusion, NAPs complement existing explainability methods with means for understanding models on a layer-level.


\bibliographystyle{abbrv-doi}

\bibliography{bibliography.bib}
\end{document}